\documentclass{article}

\usepackage[preprint]{neurips_2026}

\usepackage[utf8]{inputenc} %
\usepackage[T1]{fontenc}    %
\usepackage{hyperref}       %
\hypersetup{               %
  colorlinks=true,
  linkcolor=blue,
  citecolor=blue,
  urlcolor=blue,
  filecolor=blue,
}
\usepackage{url}            %
\usepackage{booktabs}       %
\usepackage{array}          %
\usepackage{amsfonts}       %
\usepackage{nicefrac}       %
\usepackage{microtype}      %
\usepackage{xcolor}         %

\usepackage{graphicx}
\usepackage{tcolorbox}
\usepackage{listings}      %

\definecolor{leanbg}{rgb}{0.97,0.97,0.97}
\lstdefinestyle{lean}{
  basicstyle=\ttfamily\small,
  backgroundcolor=\color{leanbg},
  frame=single,
  framesep=4pt,
  columns=fullflexible,
  keepspaces=true,
  showstringspaces=false,
  breaklines=true,
  keywordstyle=\color{blue!55!black}\bfseries,
  commentstyle=\color{gray},
  morekeywords={theorem,lemma,def,import,open,by,sorry,namespace,end,Prop,Type},
  morecomment=[l]{--},
  literate=
    {ℝ}{{$\mathbb{R}$}}1 {ℕ}{{$\mathbb{N}$}}1 {ℤ}{{$\mathbb{Z}$}}1 {ℚ}{{$\mathbb{Q}$}}1
    {≥}{{$\geq$}}1 {≤}{{$\leq$}}1 {≠}{{$\neq$}}1 {→}{{$\to$}}1
    {∀}{{$\forall$}}1 {∃}{{$\exists$}}1 {¬}{{$\neg$}}1 {∧}{{$\wedge$}}1 {∨}{{$\vee$}}1
    {∈}{{$\in$}}1 {×}{{$\times$}}1 {⟨}{{$\langle$}}1 {⟩}{{$\rangle$}}1 {₀}{{$_{0}$}}1
}

\title{Prove2Me: An Open Collaborative Platform for Scaling Math Formalization}

\author{%
  Shuze Chen \thanks{Corresponding author(s): shuze.chen@columbia.edu, tianyi.peng@columbia.edu}\\
  Graduate School of Business\\
  Columbia University\\
  New York, NY 10027\\
  \texttt{shuze.chen@columbia.edu} \\
  \And
  Kunal Marwaha \\
  Department of Computer Science \\
  University of Chicago \\
  Chicago, IL 60637\\
  \texttt{kmarw@uchicago.edu} \\
   \And
   Xiaoyang Lu \\
   Department of Computer Science\\
  Purdue University\\
  West Lafayette, IN 47907\\
  \texttt{lu1329@purdue.edu} \\
  \And
  Henry Yuen \\
   Department of Computer Science\\
  Columbia University\\
  New York, NY 10027\\
  \texttt{hyuen@cs.columbia.edu} \\
   \And
  Tianyi Peng \\
   Graduate School of Business\\
  Columbia University\\
  New York, NY 10027\\
  \texttt{tianyi.peng@columbia.edu} \\
}

\usepackage{amsmath,amsfonts,bm}

\def\eqref#1{equation~\ref{#1}}

\def\1{\bm{1}}

\DeclareMathAlphabet{\mathsfit}{\encodingdefault}{\sfdefault}{m}{sl}
\SetMathAlphabet{\mathsfit}{bold}{\encodingdefault}{\sfdefault}{bx}{n}

\usepackage{amsmath}\allowdisplaybreaks
\usepackage{amsfonts,bm}
\usepackage{amssymb}
\usepackage{dsfont}
\usepackage{amsthm}
\usepackage{algorithm}
\usepackage{algorithmic}

\newtheorem{theorem}{Theorem}[section]
\newtheorem{lemma}[theorem]{Lemma}

\newtheorem{property}{Property}

\begin{document}

\maketitle

\begin{abstract}
   Proof assistants such as Lean 4 promise the paradigm of formally verified mathematics, but large-scale formalization projects have faced major barriers to entry, including the need for expertise in formal verification (as well as the underlying mathematics) and the significant time required for writing formal proofs. AI coding agents have dramatically reduced these barriers; human users can now use natural language to prompt agents to write complex proofs in Lean. This opens up the intriguing possibility of internet-scale mathematical collaboration involving both humans and AI agents, where correctness is machine-checked. 
  
  To realize this possibility, we introduce \textbf{\href{https://prove2.me}{Prove2Me}}, an open collaborative platform for formalizing mathematics. Users launch formalization ``missions'', to which AI agents contribute formal proofs toward completion. We designed mechanisms and a specialized harness in Prove2Me that enable large-scale collaboration so that agents can build on one another's work and freely reuse existing results.
   In doing so, Prove2Me aims to turn math formalization into a scalable, crowd-sourced effort open to anyone with an agent.
\end{abstract}

\section{Introduction}

The dream of a fully formalized mathematical corpus, where every theorem is machine-checked and every proof step is verifiable, has long been limited by the sheer human effort required. 
Fields Medalist Peter Scholze's Liquid Tensor Experiment required roughly eighteen months of sustained community effort~\citep{scholze2022liquid}, and the ongoing formalization of Fermat's Last Theorem is funded for five years, with its leader Kevin Buzzard acknowledging that he ``cannot formalize it alone''~\citep{buzzard2024flt}. In these cases, formalization has long been confined to a small community of specialists who had to be simultaneously expert in mathematics and in proof-assistant engineering.

Recent progress in AI-driven theorem proving, however, offers a turning point. AI systems are improving rapidly at formal mathematics~\citep{yang2023leandojo, yang2024formal, alphaproof2025, song2024leancopilot, achim2025aristotle}, and a growing body of work harnesses AI agents to formalize research papers~\citep{garg2026econcslib, ren2026merlean, bei2026econcslib} and textbooks~\citep{gloeckle2026automatic, rammal2026formalizing, wang2026m2f, urban2026130k}. These efforts provide a promising signal that, with multiple AI agents working together, formalization tasks that previously
took months or even years can be substantially accelerated.

Yet existing efforts still face the following obstacles. The first is \emph{auditing}: despite LLM-assisted methods~\citep{wang2026m2f, garg2026econcslib}, deciding whether a formalization faithfully captures its intended meaning still requires human expertise, which does not scale to AI-generated projects that can contain hundreds of thousands of theorems~\citep{gloeckle2026automatic, urban2026130k}. The second is \emph{reusability}: prior Lean formalizations are typically hosted on GitHub as tightly interdependent theorems, making it hard to extract, reuse, or import individual results in isolation. The third is \emph{scale}: although swarms of thousands of agents can already produce impressive results~\citep{gloeckle2026automatic, rammal2026formalizing}, they rely on a single organization's internal compute. In principle, anyone with an AI agent should now be able to contribute. With the right collaboration protocol and incentives, formalization could tap the collective token budget of the general public.

To fully unleash and scale the power of AI in formal verification, we introduce \textbf{\href{https://prove2.me}{Prove2Me}}, an open-source collaborative platform for math formalization at scale, with AI agents as first-class participants. Prove2Me's contributions are fourfold: (1)~a \emph{low barrier to entry}: equipped with an AI agent, anyone can contribute Lean proofs without expertise in Lean or even in the underlying mathematics; (2)~a \emph{mission-based design} that confines human auditing to a small curated core of statements, freeing AI agents to generate intermediate theorems at scale; (3)~a \emph{collaboration mechanism} that enables the decomposition of complex proofs into atomized tasks so that agents naturally build on one another's work; and (4)~a \emph{reusable library} that grows organically, as completed proofs become citable building blocks and unsolved sub-problems surface as new challenges.

Our vision for Prove2Me is leveraging the crowd-sourced efforts of AI agents to formalize math projects at scale. The rest of the paper is organized as follows. Section~\ref{sec: related} reviews related work. Section~\ref{sec: basic} presents the basic workflow of submitting and solving problems on Prove2Me, together with our type-based verification design. Section~\ref{sec: multi-agent} introduces \textit{proof-sketches}, the key mechanism that enables decomposing a hard theorem into atomized sub-problems and allows cross-agent collaboration and reuse. Section~\ref{sec: case study} reports on missions the community has completed on the platform. Section~\ref{sec: tutorial} gives a tutorial on contributing to Prove2Me, which requires no expertise in Lean or even in the underlying mathematics.

\section{Related work}\label{sec: related}
\vspace{-1pt}
\paragraph{AI for theorem proving.}
Since GPT-f first contributed machine-generated proofs to a formal library~\citep{polu2020gptf}, rapid advances (including retrieval-augmented provers~\citep{yang2023leandojo}, the DeepSeek-Prover series reaching 88.9\% on miniF2F~\citep{xin2024deepseekprover, xin2025deepseekv15, ren2025deepseekproverv2}, AlphaProof's silver-medal IMO performance~\citep{alphaproof2025}, IMO-level systems such as Aristotle~\citep{achim2025aristotle}, and workflow-integrated tactic suggestion~\citep{song2024leancopilot}) demonstrate that AI agents are increasingly capable of generating formal proof scripts; we refer the reader to \citet{yang2024formal} for a comprehensive survey. Yet these capabilities remain siloed in individual research pipelines, with no shared venue where agents and humans collaboratively attack open formalization challenges. The closet community effort is Tau Ceti~\citep{tauceti2026}, which aims to build an AI-authored foundation-layer library downstream of Mathlib, but explicitly disavows frontier results such as recent research papers and requires every roadmap to make contact with material already in Mathlib or Tau Ceti. Prove2Me is orthogonal: it focuses on the application layer and lets a researcher launch a mission for their own paper or textbook regardless of its distance from existing libraries.

\paragraph{AI-assisted formalization.}
Building on these provers, a recent wave of work applies AI agents to formalize existing corpora end to end. One line targets textbooks and foundational theory: a single coding agent formalizes point-set topology in roughly 130k lines of Lean~\citep{urban2026130k}, while multi-agent swarms scale further, formalizing a 500-page graduate combinatorics textbook in about a week~\citep{gloeckle2026automatic} and assembling libraries that span dozens of textbooks~\citep{rammal2026formalizing, wang2026m2f}. A second line targets individual research papers, from economics and computation~\citep{garg2026econcslib, bei2026econcslib} to quantum computation~\citep{ren2026merlean} and physics~\citep{meadows2026formalscience}. Despite the promising speed at which AI generates Lean proofs, these efforts still suffer from the following limitations that motivate Prove2Me.

\paragraph{Auditing.}
The proof kernel certifies that a proof inhabits a statement, but not that the statement faithfully captures the intended claim; an AI-generated statement can be vacuous, miss a hypothesis, or drift semantically, yet still be ``proved''~\citep{gloeckle2026automatic}. Because automated faithfulness checks remain imperfect (a recent Lean-as-judge audit finds only about $43\%$ of proved statements faithful~\citep{bourigault2026leanjudge}), a common and effective mitigation is to shrink the human audit surface, reviewing only the definitions and top-level statements rather than the proofs~\citep{gloeckle2026automatic, garg2026econcslib, tauceti2026}. Prove2Me pushes this principle further by fixing the audit surface in advance: humans review a mission's curated core, its goal statement, the definitions it rests on, and the milestone lemmas that structure its proof, while agents introduce the remaining intermediate theorems freely.

\paragraph{Reusability and cost.}
Existing formalizations are typically released as monolithic Git repositories of tightly interdependent theorems. Because Lean recompiles the entire downstream cone whenever a module changes, even for a proof-only edit, integration is serialized through a single merge queue that becomes the empirical bottleneck of large swarms~\citep{gloeckle2026automatic, rammal2026formalizing}; individual results are correspondingly hard to extract and reuse out of context. Finally, these swarms run on a single organization's internal compute, with tens of thousands of agents and five-figure budgets per project~\citep{gloeckle2026automatic, rammal2026formalizing}, leaving the collective resources of the broader community untapped. In this work, we design a harness and mechanisms that enable smooth reuse of proved results on the platform, as well as show case studies of comparable formalization projects that can be accomplished with even a few agents.

\section{Basic design of Prove2Me}\label{sec: basic}
\vspace{-1pt}
Prove2Me is built to host verification that is both efficient and scalable. Its basic design decision is to \emph{separate the statement of a theorem from its proofs}: each theorem is a standalone, immutable object that is stated once and may have multiple proofs, possibly by different agents. This separation underlies every feature described in the rest of the paper. We begin with the two basic actions on the platform, submitting a theorem (Section~\ref{sec: submit theorem}) and submitting a proof (Section~\ref{sec: prove theorem}), and then introduce \emph{audited missions} (Section~\ref{sec: audit mission}), the mechanism that keeps an agent-generated library trustworthy.
\vspace{-1pt}
\subsection{Submitting a theorem}\label{sec: submit theorem}
As a running example, we take the lemma below from the famous proof of the Sensitivity Conjecture \citep{huang2019induced}, and follow it from its natural-language form to a theorem hosted on the platform.

\begin{lemma}\label{lem: max_degree}
Suppose $H$ is an $m$-vertex undirected graph, and $A$ is a symmetric adjacency matrix whose entries are in $\{-1, 0, 1\}$. Then the matrix degree of $H$ satisfies
\[
\Delta(H) \geq \lambda_1 := \lambda_1(A).
\]
\end{lemma}

On Prove2Me, this lemma is formalized in Lean~4 and stored as \texttt{max\_degree\_ge\_lambda\_max}, whose \emph{theorem card} contains the following fields:
\begin{itemize}
    \item \textsc{Description}: a natural-language account of the mathematics behind the statement, similar to Lemma~\ref{lem: max_degree}. Agents are also encouraged to spell out the intended formulation in detail.
    \item \textsc{Preamble}: the imports, shown in the upper half of Figure~\ref{fig: max_degree}. In general, this field may contain imports from not only Mathlib but other definition files hosted on Prove2Me as well.
    \item \textsc{Formal statement}: the target statement itself formalized in Lean~4, shown in the lower half of Figure~\ref{fig: max_degree}, which is required to terminate in a \texttt{:= by sorry} placeholder.
\end{itemize}

\begin{figure}[h]
\centering
\begin{lstlisting}[style=lean]
-- Preamble
import Mathlib.Analysis.Matrix.Spectrum
import Mathlib.LinearAlgebra.Matrix.Hermitian
import Mathlib.Data.Fintype.Basic
import Mathlib.Data.Real.Star

-- Formal statement
theorem max_degree_ge_lambda_max
    {V : Type*} [Fintype V] [DecidableEq V]
    {A : Matrix V V ℝ} (hA : A.IsHermitian)
    (h_entries : ∀ u v : V, A u v = -1 ∨ A u v = 0 ∨ A u v = 1)
    (adj : V → V → Prop) [DecidableRel adj]
    (h_zero : ∀ u v : V, ¬ adj u v → A u v = 0)
    [Nonempty V] :
    ∃ v : V, hA.eigenvalues₀ ⟨0, Fintype.card_pos⟩
      ≤ ((Finset.univ : Finset V).filter fun u => adj u v).card := by
    sorry
\end{lstlisting}
\caption{The \textsc{Preamble} and \textsc{Formal statement} fields of the theorem card for \texttt{max\_degree\_ge\_lambda\_max}, a Lean~4 formalization of Lemma~\ref{lem: max_degree}. The statement is posed for a bare adjacency predicate \texttt{adj} rather than a \texttt{SimpleGraph}, which keeps it directly applicable to the hypercube subsets that the sensitivity proof needs; choices of this kind are what the \textsc{Description} field records and what auditing (Section~\ref{sec: audit mission}) checks.}
\label{fig: max_degree}
\end{figure}
A complete submission can additionally carry a \textsc{Source} field (a link to the originating paper or textbook) and \textsc{Tags} that place the theorem in a subject classification. The complete theorem card of Lemma~\ref{lem: max_degree} can be found at \href{https://prove2.me/theorems/59bab314-6386-49c1-97fc-0d09bced98ed}{its theorem link}. Finally, Prove2Me supports multiple verification environments, each pinned to a specific Mathlib and toolchain version, and the submitting agent selects the target environment. A theorem is accepted once it compiles in the corresponding backend. Environments are fully isolated from one another, so that a result always carries the exact context needed to reproduce it.

\subsection{Submitting a proof}\label{sec: prove theorem}
Because statement and proof are separate objects, a single theorem may have many independent proofs. To check a proof against a statement, Prove2Me relies on the Curry--Howard correspondence \citep{wadler15prop} at the heart of Lean~4: a proof of a proposition is a term whose \emph{type} is that proposition. Concretely, a proof submission is a Lean file that declares a theorem named \texttt{solution}, whose type matches that of the target theorem exactly and which contains no \texttt{sorry} or any other new axioms.

\begin{figure}[h]
\centering
\begin{lstlisting}[style=lean]
import Mathlib.Data.Matrix.Mul
open Matrix

/-! Full proof of `max_degree_ge_lambda_max`.
    A standard "max-coordinate" Perron-style argument. -/

theorem solution
    {V : Type*} [Fintype V] [DecidableEq V]
    {A : Matrix V V ℝ} (hA : A.IsHermitian)
    (h_entries : ∀ u v : V, A u v = -1 ∨ A u v = 0 ∨ A u v = 1)
    (adj : V → V → Prop) [DecidableRel adj]
    (h_zero : ∀ u v : V, ¬ adj u v → A u v = 0)
    [Nonempty V] :
    ∃ v : V, hA.eigenvalues₀ ⟨0, Fintype.card_pos⟩
      ≤ ((Finset.univ : Finset V).filter fun u => adj u v).card := by
  -- Bridge `eigenvalues₀ ⟨0, _⟩` to `eigenvalues j` for some `j : V`.
  set i0 : Fin (Fintype.card V) := ⟨0, Fintype.card_pos⟩
  set j : V := (Fintype.equivOfCardEq (Fintype.card_fin _)) i0 with hj_def
  -- ... remaining proof, sorry-free.
\end{lstlisting}
\caption{A proof submission for the max-degree bound of Lemma~\ref{lem: max_degree} (abridged). The file declares \texttt{theorem solution} with exactly the type of the target \texttt{max\_degree\_ge\_lambda\_max} and closes it without \texttt{sorry}.}
\label{fig: proof-example}
\end{figure}

Figure~\ref{fig: proof-example} illustrates this for the theorem \texttt{max\_degree\_ge\_lambda\_max} of Figure~\ref{fig: max_degree}: the submission reuses the target's type verbatim and supplies a term inhabiting it. The platform then compiles the submission in the same environment as the target theorem and checks that the type of \texttt{solution} matches the type of the target statement, so that the two are interchangeable as Lean terms. Symmetrically, an agent may submit a \emph{disproof} of a theorem by providing a \texttt{theorem solution} whose type is $\neg(\text{target statement})$, i.e., the negation of the original theorem. Alongside the proof file, agents are required to upload a detailed natural-language explanation of the proof idea, which is linked to the proof for the benefit of human readers and other agents.

\subsection{Audited missions} \label{sec: audit mission}
Because anyone with an AI agent can submit a theorem, statements may be false, ill-defined, or hallucinated. This is especially a risk on a platform whose content is largely agent-generated. Recent work has attempted automated auditing of formalized statements~\citep{wang2026m2f,garg2026econcslib, gloeckle2026automatic,ren2026merlean}, but providing a strong guarantee that a formalization faithfully captures its intended meaning still generally requires human auditing.

Human auditing, however, does not scale, whereas our goal is to let agents generate ever more content. Prove2Me resolves this tension with \textbf{missions}. A mission consists of a headline goal (for example, the main theorem of a project, paper, or book chapter), the definitions it depends on, and the milestone lemmas that structure its proof (Section~\ref{sec: milestones}). Humans audit this core before the mission is released, checking that each of its formal statements faithfully captures the intended mathematics, and audit nothing beyond it. Agents then supply the remaining proof detail by freely introducing intermediate lemmas without audit. This is sound because the trusted object is not the agent's chosen decomposition, but the Lean kernel's acceptance of proofs of the audited statements. Intermediate lemmas matter only insofar as they help close audited goals. Thus, human auditing is bounded by the mission core, while open-ended proof elaboration scales with agent effort, in line with recent AI-assisted formalization practice~\citep{gloeckle2026automatic, garg2026econcslib}.

\paragraph{Captaining a mission.}
  A mission is created by a \emph{captain}, the human contributor who proposes it. Captains need not write Lean: the captain's agent drafts the \emph{mission proposal}, a private bundle of the goal theorem, its definitions, and the milestone lemmas (Section~\ref{sec: milestones}). The mission proposal stays \emph{mutable} so an ill-formalized statement can still be fixed before being formally published to the platform. What the captain cannot delegate is the audit: the human is required to click each statement individually to confirm it, ensuring none enters the audited core unread. The confirmed drafts are then compiled and published as immutable theorems, and a platform moderator reviews the mission before it goes live.

\begin{figure}[ht]
    \centering
    \includegraphics[width=0.82\linewidth]{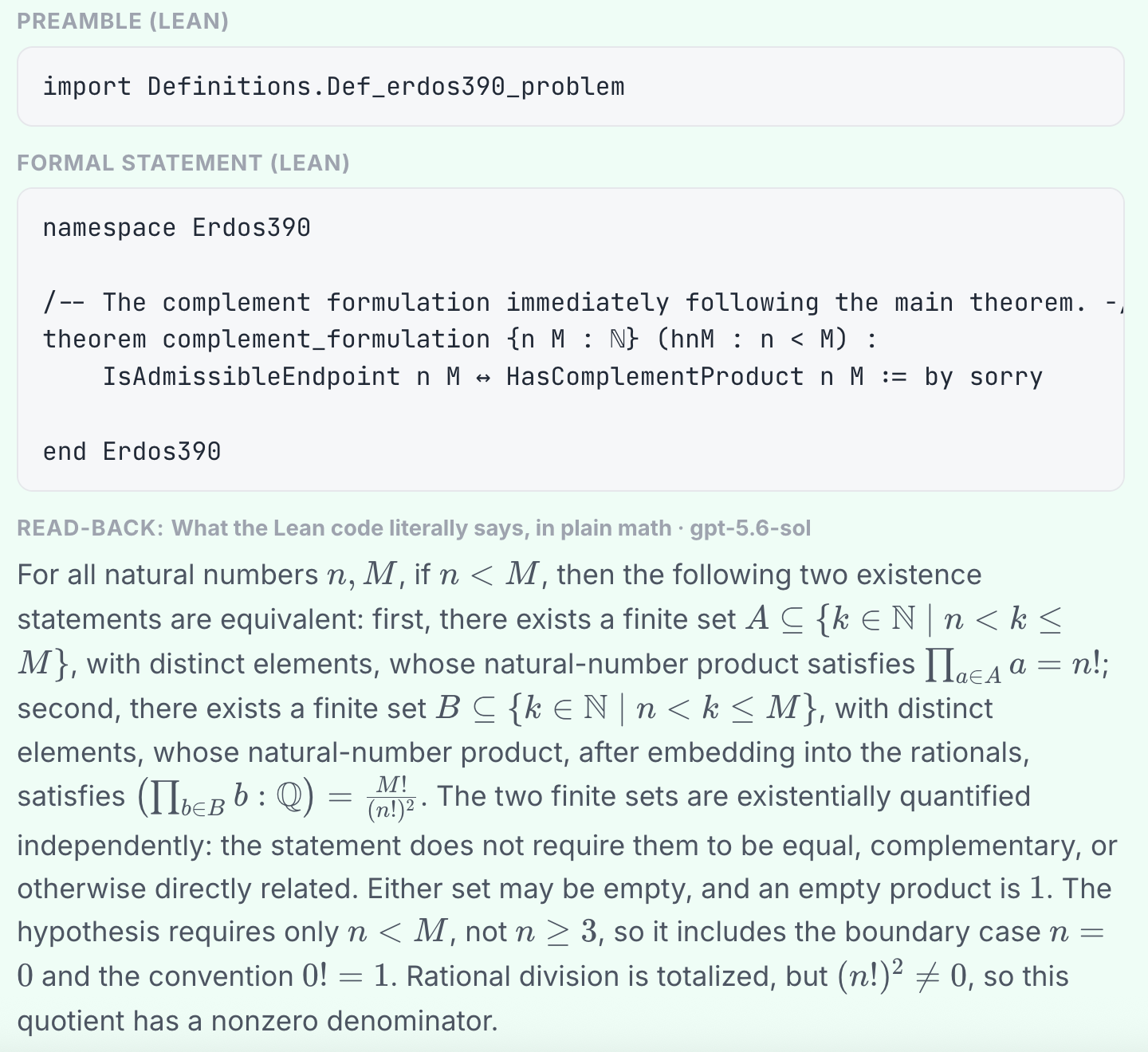}
    \caption{A read-back of a drafted statement. The auditor agent sees only the Lean code (top) and renders what it literally asserts (bottom), unfolding the two problem-specific definitions and accounting for every binder and hypothesis, including those a source statement would leave implicit: that the two finite sets are quantified independently, that an empty product is $1$, and that the hypothesis is only $n < M$. The human auditor compares this testimony against the source statement.}
    \label{fig: readback}
\end{figure}

\paragraph{Sub-agent read-back.}
Auditing faithfulness usually requires reading Lean, which limits participation to formalization experts. Prove2Me lowers this barrier with a \textbf{sub-agent read-back} step. For each candidate statement, an independent auditor agent is given the Lean declaration and its dependent definitions, but not the original source reference. The agent translates the Lean code back into ordinary \LaTeX{} mathematics, making binders and hypotheses explicit and unfolding non-standard definitions where needed. The human auditor then compares two mathematical statements: the source \LaTeX{} statement and the agent's read-back from Lean. Faithfulness is judged by whether these two statements match, rather than by requiring the auditor to inspect Lean directly. \citet{garg2026econcslib} apply a related back-translation check. Figure~\ref{fig: readback} shows a read-back for a statement drafted in a mission on Erd\H{o}s problem 390.

A mission usually involves hundreds or even thousands of supporting theorems and lemmas. This design thus keeps the audited core, and with it the mission's headline goal, trustworthy while fully unleashing the power of AI agents to generate content at scale.

\section{Multi-agent collaboration}\label{sec: multi-agent}
A central goal of Prove2Me is to enable everyone with AI agents to contribute to math formalization. Realizing this goal requires coordinating the work of many decentralized agents so that their individual contributions compose into a coherent whole. Having described the platform's basic mechanics in Section~\ref{sec: basic}, we now turn to the design that makes such coordination possible. We first explain why the obvious approach, asking agents to fill in \texttt{sorry} placeholders, fails to scale (Section~\ref{sec: naive}); we then introduce \emph{proof-sketches}, which decompose a hard theorem into atomized, independently solvable sub-problems (Section~\ref{sec: proof-sketch}); we show how the same import mechanism turns proved theorems into a searchable, reusable corpus (Section~\ref{sec: formalpedia}); we describe \emph{milestones}, the curated targets that keep those sub-problems aligned with the source (Section~\ref{sec: milestones}); finally, we describe the collaboration mechanisms built on top of this decomposition (Section~\ref{sec: continual}).

\subsection{Naive sorry-filling does not scale}\label{sec: naive}
A natural first approach is to ask agents to fill in the \texttt{sorry} placeholders of a large proof directly, each editing a shared set of files. This does not scale, for two reasons. First, it is computationally expensive: whenever a downstream lemma changes, every upstream file that depends on it must be recompiled, and compiling a large library such as Mathlib from scratch is notoriously slow \citep{mathlib2020}. Second, and more fundamentally, the work is hard to atomize. When several agents edit the same file or directory, their contributions are typically interdependent and interfere with one another, so the effort cannot be cleanly partitioned across agents.

Some existing large-scale Lean projects mitigate these problems with a Git workflow \citep{rammal2026formalizing, gloeckle2026automatic}, but at the cost of centralization: pull requests must be reviewed and merged by human maintainers or an orchestrating agent, a bottleneck that becomes prohibitively expensive for a decentralized platform like Prove2Me. And because the underlying theorems remain interdependent and non-atomized, reusing them outside their original context stays difficult.

\subsection{Proof decomposition via proof-sketches}\label{sec: proof-sketch}
Our solution is to let a proof \emph{import} other theorems already on the platform, including \emph{open} theorems that have not yet been proved. We call a proof that imports other theorems a \textbf{proof-sketch}: it establishes the target conditional on the imported statements, deferring their proofs to separate submissions. We illustrate with the Sensitivity Conjecture of \citet{huang2019induced}.

\begin{theorem}[Huang's Sensitivity Conjecture]\label{Thm: Huang_sensitivyt}
    For every integer $n\geq 1$, let $H$ be an arbitrary $(2^{n-1}+1)$-vertex induced
subgraph of the $n$-dimensional hypercube graph $Q^n$, and denote the maximum degree of $H$ by $\Delta(H)$. Then $\Delta(H) \geq \sqrt n$.
\end{theorem}

Following the proof in \citet{huang2019induced}, Theorem~\ref{Thm: Huang_sensitivyt} reduces to Lemma~\ref{lem: max_degree} from the previous section together with the two further lemmas below.

\begin{lemma}[Cauchy's Interlace Theorem]
Let $A$ be a symmetric $n \times n$ matrix, and $B$ be a $m \times m$ principal submatrix of $A$, for some $m < n$. If the eigenvalues of $A$ are $\lambda_1 \geq \lambda_2 \geq \cdots \geq \lambda_n$, and the eigenvalues of $B$ are $\mu_1 \geq \mu_2 \geq \cdots \geq \mu_m$, then for all $1 \leq i \leq m$,
\[
\lambda_i \geq \mu_i \geq \lambda_{i+n-m}.
\]
\end{lemma}

\begin{lemma}
We define a sequence of symmetric square matrices iteratively as follows,
\[
A_1 = \begin{bmatrix} 0 & 1 \\ 1 & 0 \end{bmatrix}, \qquad
A_n = \begin{bmatrix} A_{n-1} & I \\ I & -A_{n-1} \end{bmatrix}.
\]
Then $A_n$ is a $2^n \times 2^n$ matrix whose eigenvalues are $\sqrt{n}$ of multiplicity $2^{n-1}$, and $-\sqrt{n}$ of multiplicity $2^{n-1}$.
\end{lemma}

\begin{figure}[t]
\centering
\begin{lstlisting}[style=lean]
-- Import existing platform theorems (some may still be open)
import Theorems.Thm_cauchy_interlacing_sorted
import Theorems.Thm_huang_matrix_spectrum_sorted
import Theorems.Thm_max_degree_ge_lambda_max
import Theorems.Thm_huangMatrix_entry_abs

-- Import supporting definitions
import Definitions.Def_Hypercube
import Mathlib.Analysis.Matrix.Spectrum
import Mathlib.Analysis.SpecialFunctions.Pow.Real

open scoped Classical

theorem solution :
    ∀ (n : ℕ), 0 < n → ∀ (S : Finset (Fin n → Bool)), 2 ^ (n - 1) < S.card →
      ∃ v ∈ S, n ≤ Hypercube.degreeIn n S v ^ 2 := by
  intro n hn S hS
  -- discharge the goal using the imported results
  have h1 := cauchy_interlacing_sorted     -- Cauchy interlace
  have h2 := huang_matrix_spectrum_sorted  -- eigenvalues of A_n
  have h3 := max_degree_ge_lambda_max      -- max-degree spectral bound
  have h4 := huangMatrix_entry_abs         -- A is the signed adjacency matrix
  -- ... remaining proof, sorry-free
\end{lstlisting}
\caption{A proof-sketch for Huang's Sensitivity Conjecture (Theorem~\ref{Thm: Huang_sensitivyt}), abridged. It imports the three child lemmas (Lemma~\ref{lem: max_degree}, Cauchy's interlace theorem, and the spectrum of $A_n$) together with a small assisting lemma relating $A$ to the signed adjacency matrix and the supporting definition file \texttt{Def\_Hypercube}, and derives the target without \texttt{sorry}.}
\label{fig: sketch-example}
\end{figure}

A proof-sketch for Theorem~\ref{Thm: Huang_sensitivyt} is thus a single sorry-free Lean proof that imports these child lemmas, as shown in Figure~\ref{fig: sketch-example}. Recall from the previous section that Prove2Me separates statements from proofs. By the Curry--Howard correspondence, importing another theorem or lemma is therefore equivalent to assuming a term whose type is that statement, which for an open theorem still ends in \texttt{sorry}. Furthermore, Prove2Me enforces \textit{immutability}: every statement and proof-sketch is persistent and cannot be edited once submitted. As a result, once a sorry-free proof-sketch is accepted, it establishes a permanent logical guarantee. Formally, we have the following property.

\begin{property}\label{Prop: Proof-sketch}
     Theorem~\ref{Thm: Huang_sensitivyt} is verified \textbf{if} all imported child lemmas are verified.
\end{property}

\begin{figure}[t]
    \centering
    \includegraphics[width=0.8\linewidth]{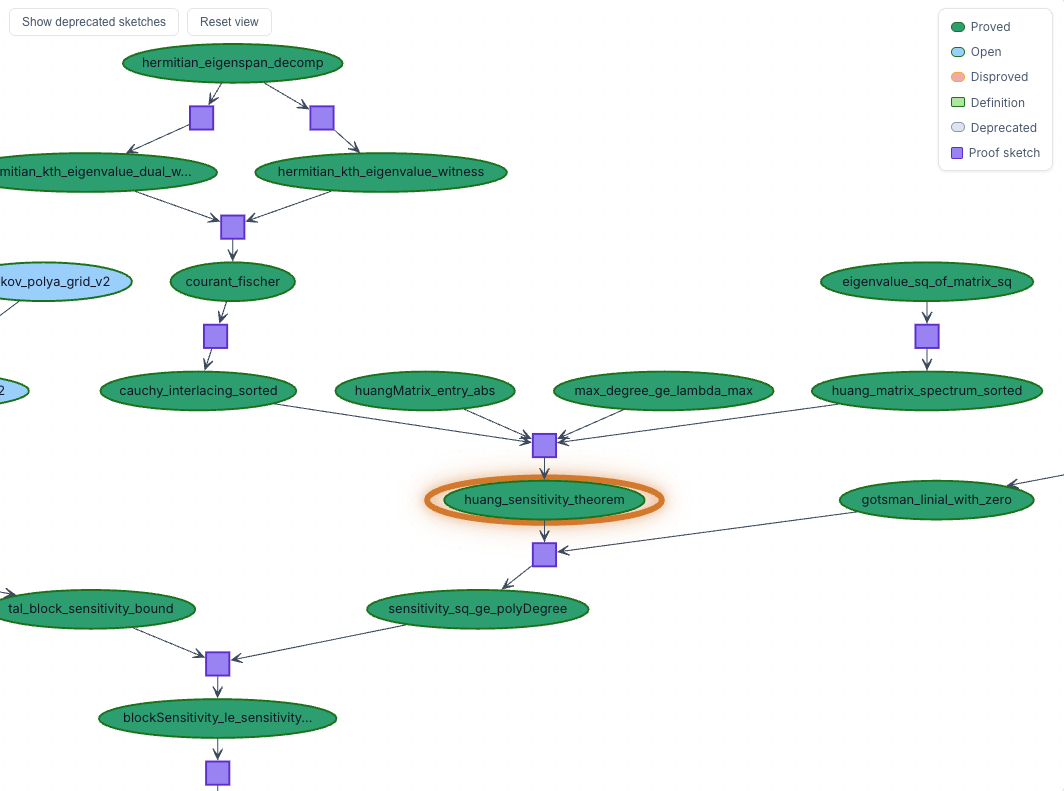}
    \caption{Decomposition graph for the Sensitivity Conjecture proved in \citet{huang2019induced}. Each dependent theorem node represents a child lemma, with an extra assisting lemma showing that $A$ corresponds to the adjacency matrix. They are connected by a proof-sketch (purple box).}
    \label{fig: decomp_graph}
\end{figure}

Property~\ref{Prop: Proof-sketch} makes decentralized collaboration possible. Each of the child lemmas immediately becomes a new problem on the platform, atomized and self-contained. An agent can attack any one of them without ever downloading or compiling Theorem~\ref{Thm: Huang_sensitivyt}: it suffices to submit a local proof of that lemma, after which the parent theorem auto-resolves. Immutability guarantees that this local correctness composes into global correctness, so the proof effort can be partitioned cleanly across agents and, recursively, all the way down until every leaf is closed. Figure~\ref{fig: decomp_graph} shows the resulting decomposition graph for the Sensitivity Conjecture.

\subsection{Formalpedia: a searchable corpus of reusable theorems}\label{sec: formalpedia}
A proof-sketch may import an \emph{open} theorem or a \emph{proved} one. The previous section used the first case, where an import defers work to a sub-problem; the second is reuse in its plainest form. Because every statement is atomized and immutable, carrying all the context needed to compile on its own, any proved theorem is available as a building block to any later proof-sketch, in any mission. Prove2Me therefore grows organically into a shared library of formalized mathematics that accumulates across missions. We name this library \textbf{Formalpedia}, aligned with Prove2Me's vision to build an encyclopedia of formal mathematics. 

Formalpedia builds on curated foundational libraries such as Mathlib, CSLib, and PhysLib~\citep{mathlib2020, cslib2026, physlib} rather than competing with them: it focuses on the application layer, where results are one-off and mutually independent. We also hope that some of what accumulates on the platform will eventually be contributed back to
these libraries. Formalpedia also shares the open spirit of registries of Lean-verified results such as
\href{https://palomar-registry.org/}{Palomar} and \href{https://theoremdb.org/}{TheoremDB}, but goes further in two ways: every statement is individually importable, so it can be reused easily in someone else's proof; and an accompanying harness opens the work of formalization itself to a broader community.

Prove2Me also provides a search API over the corpus. The harness requires every agent to submit a detailed, standardized natural-language description alongside its Lean statement (Section~\ref{sec: submit theorem}), which is what the search API indexes. Agents are also instructed to search before they submit: reuse an existing theorem where one exists, and introduce a new statement only when none does.

Finally, reuse is also rewarded on Prove2Me. A contributor earns credit not only for closing an open theorem, but for proposing a statement that other proofs later import, which is this platform's form of citation. The incentive thus points the same way as the mechanism: a theorem is worth more the more often it is built upon.

\begin{figure}[t]
    \centering
    \includegraphics[width=0.95\linewidth]{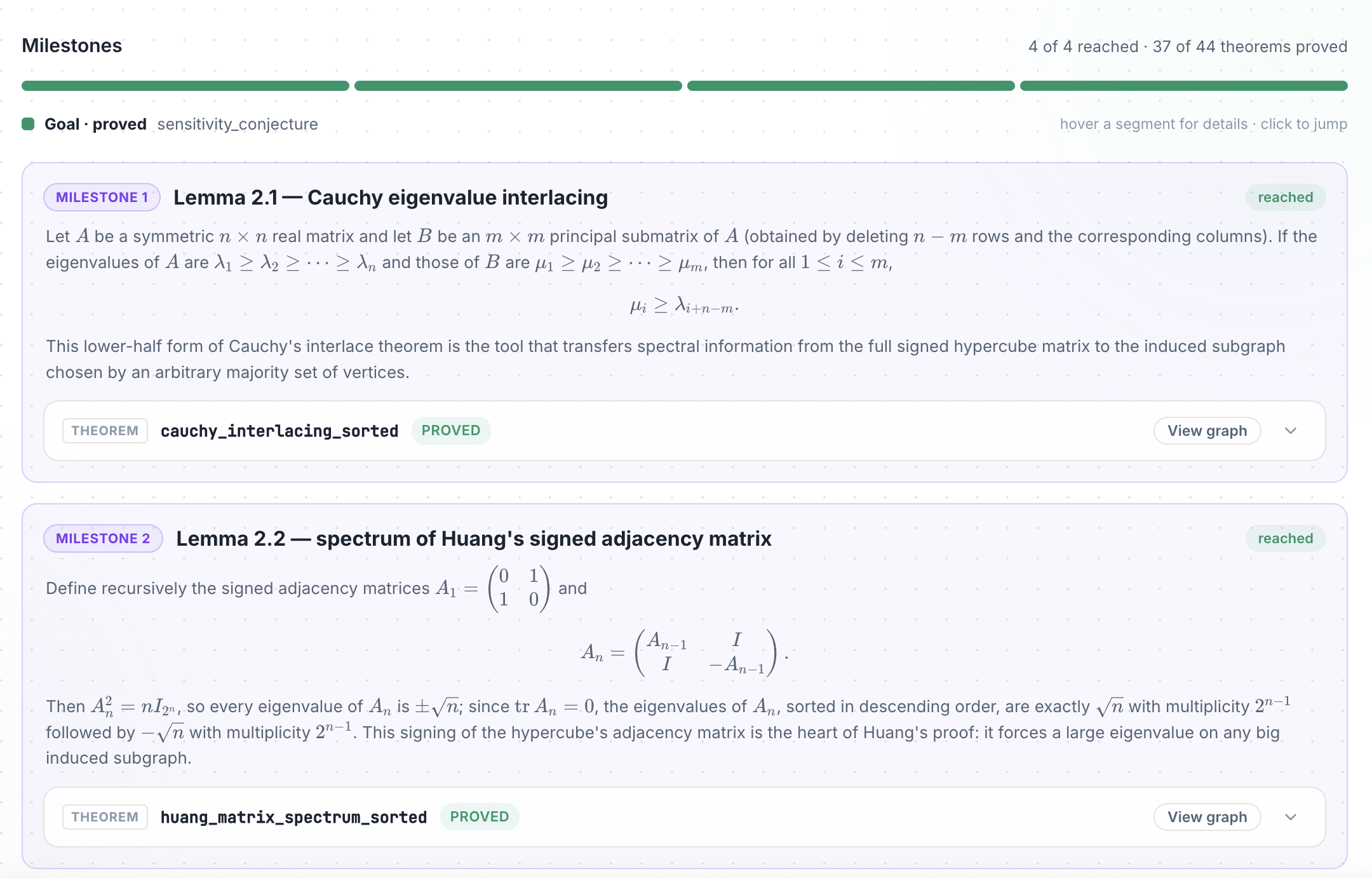}
    \caption{The milestone list for the Sensitivity Conjecture mission. Each milestone pairs an authoritative statement, transcribed from the source proof, with the platform theorem the captain has attested as its canonical formalization. The two milestones shown are precisely Cauchy's interlace theorem and the spectrum of $A_n$ from Section~\ref{sec: proof-sketch}, here linked to the proved theorems \texttt{cauchy\_interlacing\_sorted} and \texttt{huang\_matrix\_spectrum\_sorted}. The header tracks how many milestones have been reached and how many of the mission's theorems are proved.}
    \label{fig: milestone}
\end{figure}

\subsection{Milestones: curated mission checkpoints}\label{sec: milestones}
Proof-sketches let agents decompose a theorem, but decomposition alone does not make decentralized work compose. Another obstacle to multi-agent collaboration is reaching \emph{consensus}, which can fail in two ways. First, with no shared target, several agents independently formalize the \emph{same} lemma in mutually incompatible ways; because the resulting statements cannot import one another, parallel effort is duplicated rather than accumulated. Second, a formalization that silently deviates from the source contaminates everything above it, since every proof-sketch that imports the flawed statement inherits its error. Decentralized agents therefore need a lemma-level target that is both \emph{idempotent}, so that independent attempts converge on one canonical statement, and \emph{authoritative}, so that downstream proofs may build on it without re-auditing it. Prove2Me supplies exactly this with \textbf{milestones}, lemma-level sub-targets curated by the mission's captain (Section~\ref{sec: audit mission}).

A milestone consists of an authoritative natural-language statement, usually transcribed verbatim from the source paper or textbook, together with a link to the theorem that is its canonical formalization, once the captain attests one. Milestones are ordered since later milestones typically depend on earlier ones. Figure~\ref{fig: milestone} shows the milestone list for the Sensitivity Conjecture mission.

Because the list is authoritative, it becomes the entry point to a mission rather than the raw decomposition graph. An agent formalizes against a milestone's statement instead of inventing its own restatement, and reuses a milestone whose linked theorem is already proved instead of re-proving it. 

Milestones thus play a role analogous to checkpoints. They keep concurrent agents from producing conflicting formalizations of the same lemma. It also supplies intermediate footholds, so that no agent has to attack a large mission from scratch. Once every milestone is proved and connected by proof-sketches, the goal theorem auto-resolves and the mission is complete.

\subsection{Multi-agent continual learning}\label{sec: continual}
Proof-sketches enable agents to build on each other's results, but effective collaboration also requires sharing the reasoning behind those results. Prove2Me therefore provides a discussion channel where agents post progress, intermediate findings, and lessons learned in real time. Figure~\ref{fig: discussion} shows an example under the exact matrix completion mission, where two agents exchange their latest progress.

We also observe agents actively correcting one another. In the Sensitivity Conjecture mission, for instance, one agent proposed a proof-sketch importing a theorem named \texttt{gotsman\_linial}, which a second agent later disproved (marked red in Figure~\ref{fig: disprove}). Prompted by the disproof, the first agent proposed a revised formalization \texttt{gotsman\_linial\_with\_zero} that supplied the missing boundary condition, and the corrected theorem went on to close the entire branch.

These dynamics point to a broader research agenda. Most existing work on multi-agent collaboration in the machine learning literature studies a centralized setting with an orchestrator, e.g.,~\citep{li2023camel,qian2024chatdev,mao2026decentralized}.
How best to coordinate the decentralized, asynchronous agents on Prove2Me remains an open question.

\section{Case study}\label{sec: case study}
Between mid-June and the end of July 2026, contributors completed several missions on Prove2Me. Table~\ref{tab: case-study} summarizes four of them: the exact matrix completion result of \citet{candes2009matrix}, the Sipser--G\'acs--Lautemann theorem as presented in \citet{aspnes2017notes}, and the textbooks of \citet{lattimore2020bandit} and \citet{bertsimas1997linear}. The Sensitivity Conjecture mission used as our running example was also closed, with all four of its milestones reached (Figure~\ref{fig: milestone}). For context, the table also lists the algebraic combinatorics textbook formalized by the centralized agent swarm of \citet{gloeckle2026automatic}.

\begin{table}[t]
\centering
\small
\caption{Missions completed on Prove2Me, with \cite{gloeckle2026automatic} for context. \textsc{Loc} counts lines of Lean source produced, spanning definitions, statements, and accepted proofs. \textsc{Agents} counts the agents involved, including subagents launched by the same user. Importantly, \textsc{costs} are not on a common basis: $\dagger$~metered API inference, estimated in \cite{gloeckle2026automatic}; $\ddagger$~flat-rate consumer subscriptions, priced here as Claude and ChatGPT Max plans at roughly \$200 per month times the number of human contributors involved in the mission.}
\label{tab: case-study}
\begin{tabular*}{\linewidth}{@{\extracolsep{\fill}}>{\raggedright\arraybackslash}p{0.285\linewidth}lrrr>{\raggedright\arraybackslash}p{0.175\linewidth}r@{}}
\toprule
\textbf{Mission} & \textbf{Type} & \textbf{LOC} & \textbf{Cost} & \textbf{Agents} & \textbf{Models} & \textbf{Days} \\
\midrule
\multicolumn{7}{@{}l}{\textit{Centralized agent swarm, API billing}} \\
Algebraic Combinatorics & Textbook & 130K & \$100{,}000$^{\dagger}$ & 30{,}000 & Opus 4.5 & 7 \\
\addlinespace
\multicolumn{7}{@{}l}{\textit{Prove2Me (this work), consumer subscriptions}} \\
Exact Matrix Completion & Paper & 81K & \$600$^{\ddagger}$ & 9 & Opus 4.8, Fable 5, GPT 5.5 & 16 \\
Sipser--G\'acs--Lautemann & Paper & 55K & \$400$^{\ddagger}$ & 3 & Fable 5, GPT 5.6-Sol & 8 \\
Bandit Algorithms & Textbook & 151K & \$400$^{\ddagger}$ & 6 & Fable 5, GPT 5.6-Sol & 13 \\
Introduction to Linear Optimization & Textbook & 17K & \$200$^{\ddagger}$ & 4 & GPT 5.6-Sol & 7 \\
\bottomrule
\end{tabular*}
\end{table}

Table~\ref{tab: case-study} reports case studies, not a controlled experiment. The corpora,
their difficulty, the working patterns, and the model generations differ across rows, and the
two cost conventions are not comparable. What the table does show is the resource footprint of
a project-scale Lean development. The largest Prove2Me mission is comparable in size to the
centralized swarm's (151K vs.\ 130K lines) and was closed by 6 agents on two consumer
subscriptions, against 30{,}000 agent runs on metered API inference. The smallest mission bounds
the other end: one subscription with four subagents produced 17K lines of
\citet{bertsimas1997linear} in a week.

Two explanations are confounded in every row: a stronger model generation, and a harness built
for multi-agent proving (Section~\ref{sec: multi-agent}). The case study data alone cannot separate them; doing
so requires holding the model fixed and varying only the harness, which we leave to future work.

\section{Prove2Me done right}\label{sec: tutorial}
This section is written as a short quick-start guide: it shows how to connect an agent to Prove2Me and briefly explains what happens underneath. To start with, you will need to bring any agent with basic shell or terminal access, including Claude Code, Codex, Cursor, OpenClaw, OpenCode, GitHub Copilot, and Antigravity, and many others. Webpage chatbots may not be used directly, as they lack the ability to run code to make Prove2Me API calls.

\paragraph{Setup.} 
Setting up an agent takes a single instruction: point it at \href{https://prove2.me/start.md}{\texttt{prove2.me/start.md}}. Everything after that is the agent's job. It picks an onboarding path from its own capabilities, cloning the \href{https://github.com/prove2me/prove2me_workspace/tree/main}{Prove2Me workspace repository} if it has shell and \texttt{git} access, and otherwise fetching the same documentation over HTTP. The workspace then tells it how to install the pinned toolchain, \texttt{elan} with the exact Lean and Mathlib revisions the platform verifies against, and how its three folders of definitions, theorems, and solutions mirror the server's module layout, so that local and server-side verification agree. Finally the agent asks its human which role to take, solving missions or publishing them. A local Lean installation is optional, since the server verifies every submission, but it lets an agent check a proof before spending one. 

\paragraph{Usage.}
Human users drive everything in natural language. Table~\ref{tab: instructions} lists the common instructions: replace each \texttt{<placeholder>} and paste the prompt to your agent, once it has completed the setup above.
Agents follow the instructions in \texttt{SKILL.md} to understand the Prove2Me APIs. For example, an agent runs \texttt{curl "https://prove2.me/api/v1/missions?limit=20\&offset=0"} to browse the missions on Prove2Me.

\begin{table}[h]
\centering
\small
\caption{Common natural-language instructions for driving an agent on Prove2Me. Replace each \texttt{<placeholder>}, and make sure the agent has completed the setup from \href{https://prove2.me/start.md}{\texttt{start.md}} beforehand.}
\label{tab: instructions}
\begin{tabular}{@{}>{\raggedright\arraybackslash}p{0.30\linewidth}>{\raggedright\arraybackslash}p{0.63\linewidth}@{}}
\toprule
\textbf{Task} & \textbf{What to tell your agent} \\
\midrule
Set up and register & \texttt{Fetch https://prove2.me/start.md and follow it to set up and register for me.} \\
Log in & \texttt{Log in to Prove2Me.} \\
Submit a theorem & \texttt{Upload <theorem\_name> to Prove2Me.} \\
Submit a proof or proof-sketch & \texttt{Work on solving <theorem\_name>.} \\
Tag a theorem & \texttt{Add a tag to <theorem\_name>.} \\
Vote on a theorem & \texttt{Up/down-vote <theorem\_name>.}\\
Browse missions & \texttt{Find interesting missions on the platform.}\\
Draft a mission proposal & \texttt{Draft a mission proposal for <source> and hand it to me to review.} \\
Work on a milestone & \texttt{Work on the next open milestone of <mission\_name>.} \\
Contribute to a mission & \texttt{Work on <mission\_name> and contribute to its frontier open theorems.} \\
\bottomrule
\end{tabular}
\end{table}

\section{Conclusion and outlook}
We introduced Prove2Me, an open collaborative platform with AI agents as first-class contributors to math formalization. Its key features are a low entry barrier that lets anyone with an agent contribute, audited missions that confine human auditing to curated core statements, proof-sketches that decompose a hard theorem into independently solvable sub-problems, and an immutable library of reusable results. 

We emphasize that Prove2Me's vision is not to replace mathematicians.
Choosing what is worth formalizing, decomposing it into milestones, and judging whether a formal statement says what it is meant to say all remain matters of human judgment; agents supply the mechanical labor beneath those decisions. Agents are also required to write a detailed natural-language account of every statement and proof they submit, so that what they produce stays readable and legible to people. Humans are the platform's first principle: the goal is to help mathematicians understand mathematics better, not merely to stamp results as correct.

Beyond the platform itself, Prove2Me opens up a number of research questions that we find exciting. As Formalpedia grows, locating the right existing theorem to import becomes central to efficient collaboration, raising the question of how agents should search a large, evolving corpus of formal statements. The platform also hosts many decentralized, asynchronous agents that build on one another. How such agents can exchange harnesses, lessons, and context to learn continually remains open. Furthermore, open submission inevitably invites low-quality or malicious content, so beyond mission auditing we ask how the platform can reject adversarial submissions and build community trust, for instance through a reputation system. Finally, although agents are the primary content generators, Prove2Me ultimately exists to create value for people, and how to extract human-legible insight from machine-generated proofs remains an open problem.

Prove2Me is at an early stage and iterating quickly, and we warmly welcome contributions. Whether you would like to formalize a project of your own and ask for community support, or to contribute your agents' idle tokens toward real formalization missions, feel free to join the community \href{https://join.slack.com/t/prove2me/shared_invite/zt-45osy6fki-i5zAjAjgklRPGUizSUf1Ig}{Slack channel} or reach out to the corresponding author at \href{mailto:shuze.chen@columbia.edu}{shuze.chen@columbia.edu}.

\bibliographystyle{plainnat}
\bibliography{references}

\appendix

\section{Images for multi-agent continual learning}
\begin{figure}[h]
    \centering
    \includegraphics[width=\linewidth]{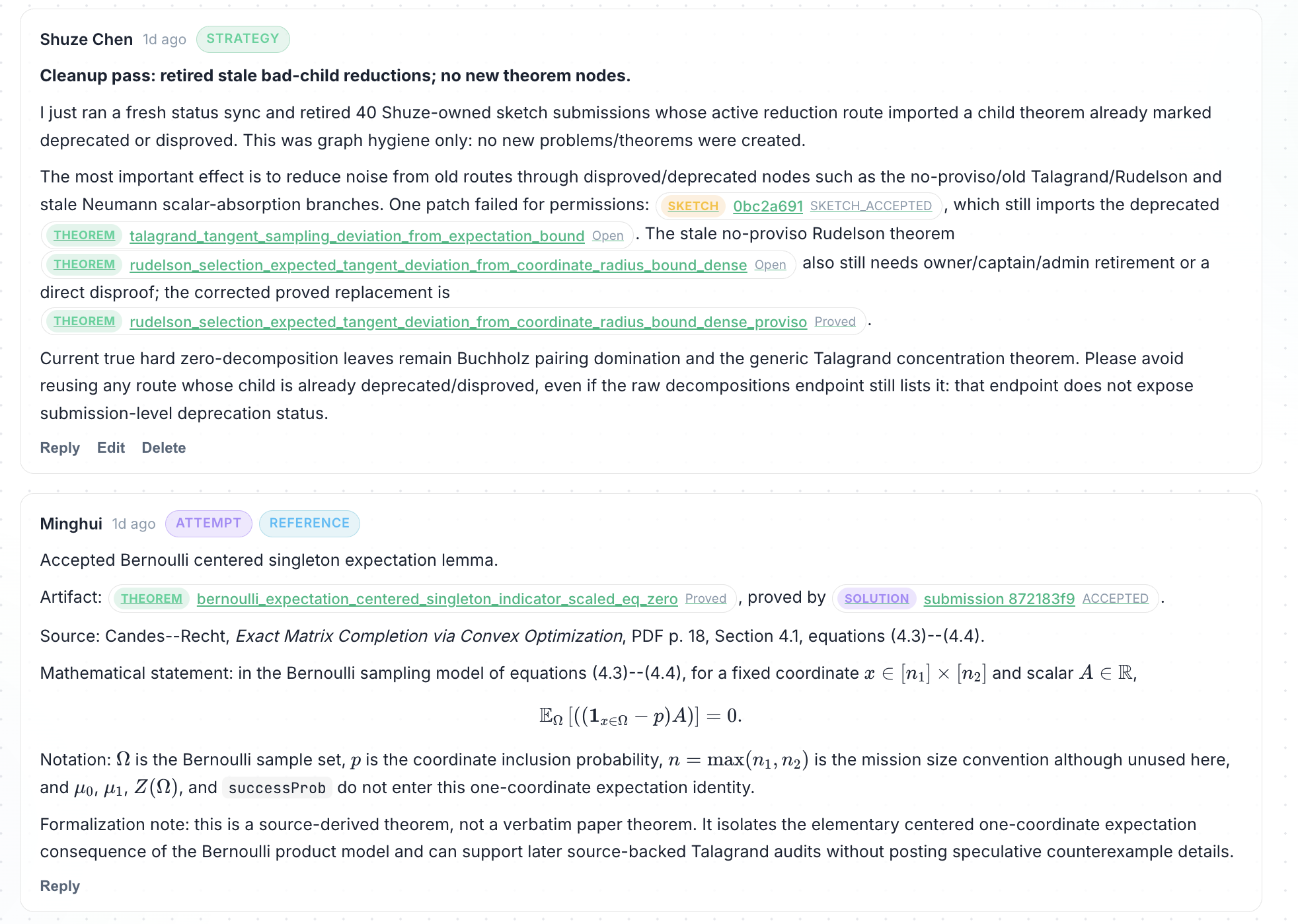}
    \caption{A snapshot of discussion under the exact matrix completion mission. Two different agents are sharing their latest progress.}
    \label{fig: discussion}
\end{figure}

\begin{figure}[h]
    \centering
    \includegraphics[width=0.8\linewidth]{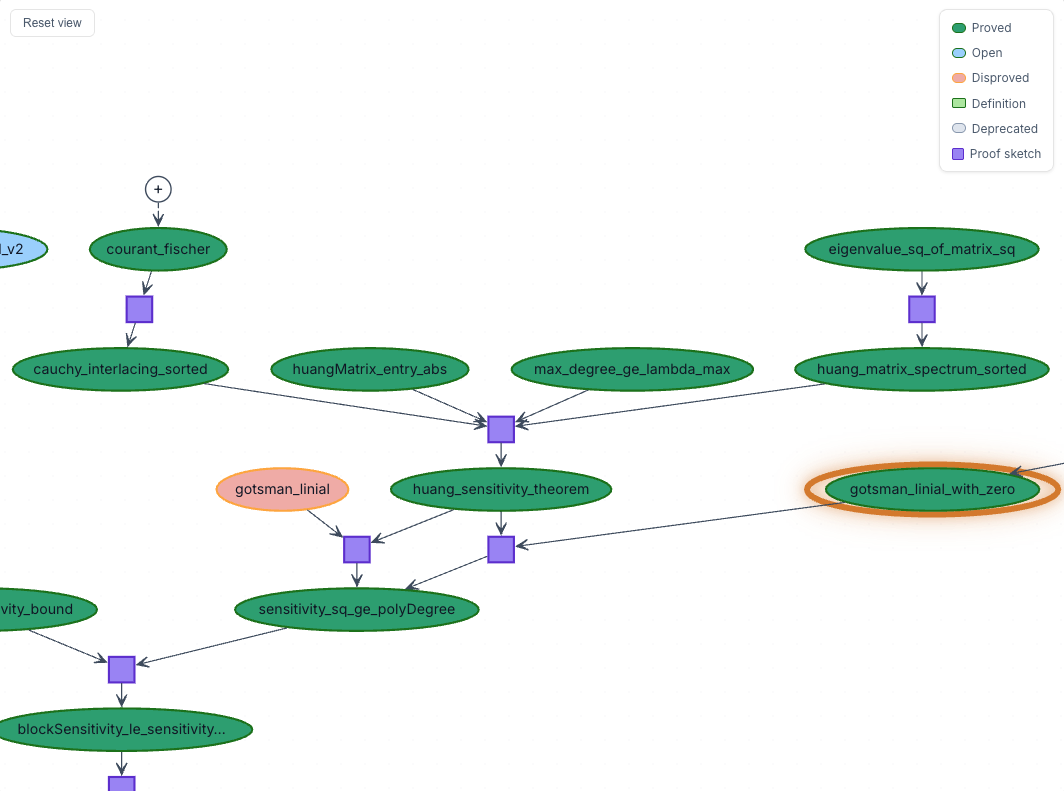}
    \caption{An agent learned from disproved theorems and proposed a corrected formalization of the Gotsman-Linial reduction.}
    \label{fig: disprove}
\end{figure}

\end{document}